\documentclass[10pt,letterpaper,twocolumn]{article}
\usepackage[
  letterpaper,
  top=0.75in,
  bottom=1in,
  left=0.75in,
  right=0.75in
]{geometry}
\usepackage{graphicx}
\usepackage{array}
\usepackage{fontspec}
\usepackage{listings}
\usepackage{fvextra}
\usepackage{microtype}
\usepackage{pgfplots}
\usepackage{titlesec}
\usepackage{tcolorbox}
\usepackage{xurl}
\usepackage{tikz}

\definecolor{customorange}{HTML}{FF7F00}

\usepackage{xcolor}
\usepackage{hyperref}
\hypersetup{
    colorlinks=true,
    citecolor=customorange,
    urlcolor=customorange,
    linkcolor=black
}
\usepackage{amssymb}
\usepackage[hang,flushmargin]{footmisc}
\usetikzlibrary{patterns.meta}

\usepackage{caption}
\usepackage{float}
\newcommand{\figref}[1]{\hyperref[#1]{\textcolor{customorange}{Fig.~\ref*{#1}}}}
\newcommand{\appendixref}[1]{\hyperref[#1]{\textcolor{customorange}{Appendix~\ref*{#1}}}}

\usepackage[outline]{contour}
\contourlength{0.8pt}

\definecolor{sectionUnderlineColor}{HTML}{6a00ff}

\newcommand{\COMMENT}[1]{}

\pgfplotsset{compat=1.16}
\definecolor{relayred}{HTML}{E31A1C}
\definecolor{hleblue}{HTML}{BDD7E7}
\definecolor{arcblue}{HTML}{6BAED6}
\definecolor{taublue}{HTML}{2171B5}
\definecolor{anthropicorange}{HTML}{D97757}
\definecolor{openaigreen}{HTML}{10A37F}
\definecolor{googleblue}{HTML}{9334E6}
\definecolor{deepseekblue}{HTML}{4D6BFE}
\definecolor{indigo}{HTML}{4F46E5}
\definecolor{turquoise}{HTML}{14B8A6}
\colorlet{gtl}{green!60!white}
\pgfdeclareverticalshading{tlgreenshade}{100bp}{color(0bp)=(white); color(58bp)=(white); color(100bp)=(gtl)}

\tcbuselibrary{breakable,skins}

\defaultfontfeatures{Ligatures=TeX}
\titleformat{\section}[hang]
  {\large\bfseries}
  {\thesection}
  {0.75em}
  {\MakeUppercase}
\titlespacing*{\section}{0pt}{1.25\baselineskip}{0.55\baselineskip}

\titleformat{\subsection}[hang]
  {\normalsize\bfseries}
  {\thesubsection}
  {0.75em}
  {}

\titlespacing*{\subsection}{0pt}{1\baselineskip}{0.5\baselineskip}

\newfontfamily\FONTjetbrainsmono{JetBrainsMono-Regular.ttf}[Scale=0.85]
\newfontfamily\FONTjetbrainsmonobold{JetBrainsMono-ExtraBold.ttf}[Scale=0.85]

\newcommand{\relaybench}{\texorpdfstring{{\textit{Relay-Bench}}}{Relay-Bench}}

\fvset{
  breaklines=true,
  breakanywhere=true,
  breaksymbolleft={},
  breaksymbolright={},
  breakanywheresymbolpre={},
  breakanywheresymbolpost={},
  breakindent=0em,
  fontsize=\small,
  formatcom=\FONTjetbrainsmono,
}

\newtcolorbox{promptbox}{
  breakable,
  enhanced,
  colback=black!1,
  colframe=black!45,
  boxrule=0.95pt,
  arc=2.2mm,
  left=3mm, right=3mm, top=2mm, bottom=2mm,
  overlay first={
    \draw[line width=1.4pt, white]
      ([xshift=2mm]frame.south west) -- ([xshift=-2mm]frame.south east);
    \draw[dashed, line width=0.95pt, black!45]
      ([xshift=2mm]frame.south west) -- ([xshift=-2mm]frame.south east);
  },
  overlay middle={
    \draw[line width=1.4pt, white]
      ([xshift=2mm]frame.north west) -- ([xshift=-2mm]frame.north east);
    \draw[dashed, line width=0.95pt, black!45]
      ([xshift=2mm]frame.north west) -- ([xshift=-2mm]frame.north east);
    \draw[line width=1.4pt, white]
      ([xshift=2mm]frame.south west) -- ([xshift=-2mm]frame.south east);
    \draw[dashed, line width=0.95pt, black!45]
      ([xshift=2mm]frame.south west) -- ([xshift=-2mm]frame.south east);
  },
  overlay last={
    \draw[line width=1.4pt, white]
      ([xshift=2mm]frame.north west) -- ([xshift=-2mm]frame.north east);
    \draw[dashed, line width=0.95pt, black!45]
      ([xshift=2mm]frame.north west) -- ([xshift=-2mm]frame.north east);
  },
}

\newlength{\figonelegendgap}
\newlength{\figtwotoplift}
\newcommand{\topfigureblock}{%
\noindent
\begin{minipage}[t]{\columnwidth}
\vspace{0pt}%
\vspace{-7pt}%
\centering
\resizebox{\linewidth}{!}{%
\begin{tikzpicture}[>=stealth, align=center, thick, font=\normalsize, baseline=(current bounding box.north)]
\begin{scope}[rotate=90, transform shape]
\tikzset{
    fig1textdown/.style={transform canvas={yshift=-1pt}},
    edge_arrow/.style={->, line width=1.2pt, draw=gray},
    base_node/.style={circle, minimum size=0.4cm, inner sep=0pt},
    encoded/.style={base_node, draw=black, line width=1pt, preaction={fill=black}, pattern={Lines[angle=45,distance=2pt,line width=1pt]}, pattern color=white},
    decoded/.style={base_node, draw=black, line width=1.5pt, fill=black},
    final/.style={base_node, draw=black, line width=1.5pt, fill=gray!40},
    context/.style={base_node, draw=black, line width=1.5pt, fill=white},
    IE/.style={base_node, draw=black, dashed, line width=1.5pt, fill=blue!25},
    MA/.style={base_node, draw=black, dashed, line width=1.5pt, fill=red!40},
    DA/.style={base_node, draw=black, dashed, line width=1.5pt, fill=green!20},
    GK/.style={base_node, draw=black, dashed, line width=1.5pt, fill=purple!45}
}

\draw[dashed, line width=1.5pt, gray]
    ([xshift=15pt]-4.5, 0) -- ([xshift=2pt]4.5, 0);
\node[fig1textdown, font=\normalsize\bfseries, text=black, rotate=-90, align=right, inner sep=0pt, anchor=north east]
    at (4.5, 0.22) {Problem\\formulation};
\node[fig1textdown, font=\normalsize\bfseries, text=black, rotate=-90, align=left, inner sep=0pt, anchor=north west]
    at (4.5, -0.22) {Solution\\formulation};

\node[decoded] (dec_left)  at (-1.1,  2.2) {};
\node[context] (bloat)     at ( 1.1,  2.2) {};
\node[encoded] (enc)       at ( 0.0,  0.0) {};
\node[decoded] (dec_right) at ( 0.0, -2.2) {};

\node[IE] (sp8)  at ( 3.6, -4.4) {};
\node[GK] (sp4)  at ( 3.0, -4.4) {};
\node[GK] (sp5)  at ( 2.4, -4.4) {};
\node[IE] (sp6)  at ( 1.8, -4.4) {};
\node[IE] (sp7)  at ( 1.2, -4.4) {};
\node[IE] (sp1)  at ( 0.6, -4.4) {};
\node[DA] (sp2)  at (-0.6, -4.4) {};
\node[GK] (sp3)  at (-1.3, -4.4) {};
\node[IE] (sp9)  at (-2.0, -4.4) {};
\node[IE] (sp10) at (-2.6, -4.4) {};
\node[GK] (sp11) at (-3.2, -4.4) {};
\node[GK] (sp12) at (-3.8, -4.4) {};

\node[MA]    (integral) at ( 0.0, -6.3) {};
\node[final] (final)    at ( 0.0, -8.2) {};

\draw[edge_arrow] (bloat) to[out=270, in=90] (enc);

\draw[edge_arrow] (dec_left) to[out=270, in=90]
    node[fig1textdown, font=\normalsize, midway, right=3pt, xshift=-6pt, rotate=-160] {Augmentation}
    (enc);

\draw[edge_arrow] (enc.south) to[out=270, in=90]
    node[fig1textdown, font=\normalsize, midway, right=3pt, xshift=-6pt, rotate=-160] {Reversal of augmentation}
    (dec_right.north);

\foreach \i in {1,2,3,4,5,6,7,8,9,10,11,12} {
    \draw[edge_arrow] (dec_right) to[out=270, in=90] (sp\i);
}

\draw[edge_arrow] (sp1) to[out=270, in=90] (integral);
\draw[edge_arrow] (sp2) to[out=270, in=90] (integral);
\draw[edge_arrow] (sp3) to[out=270, in=90] (integral);

\draw[edge_arrow] (dec_right) to[out=270, in=90] (integral);

\draw[edge_arrow] (integral.south) to[out=270, in=90] (final.north);

\foreach \i in {4,5,6,7,8,9,10,11,12} {
    \draw[edge_arrow] (sp\i) to[out=270, in=90] (final);
}

\end{scope}
\end{tikzpicture}%
}
\par\vskip\figonelegendgap
\noindent\begin{minipage}{\linewidth}
\centering
\resizebox{\linewidth}{!}{%
\begin{tikzpicture}[x=0.9cm, y=0.5cm, font=\small, >=stealth]
    \tikzset{
        leg_node/.style={circle, minimum size=10pt, inner sep=0pt},
        l_enc/.style={leg_node, draw=black, line width=1pt,
                      preaction={fill=black},
                      pattern={Lines[angle=45,distance=2pt,line width=1pt]},
                      pattern color=white},
        l_dec/.style={leg_node, draw=black, line width=1.2pt, fill=black},
        l_fin/.style={leg_node, draw=black, line width=1.2pt, fill=gray!40},
        l_bloat/.style={leg_node, draw=black, line width=1.2pt, fill=white},
        l_sub/.style={leg_node, draw=black, dashed, line width=1.2pt, fill=white},
        l_ie/.style={leg_node, fill=blue!25},
        l_ma/.style={leg_node, fill=red!40},
        l_da/.style={leg_node, fill=green!20},
        l_gk/.style={leg_node, fill=purple!45},
        leg_text/.style={anchor=mid west, font=\small, transform canvas={yshift=-1pt}},
        leg_line/.style={line width=1.2pt, draw=gray}
    }
    \node[l_bloat] (legbloat) at (0.2, 1.0) {};
    \node[leg_text] at (0.48, 1.0) {Context bloat};
    \node[l_dec] at (0.2, 0.0) {};
    \node[leg_text] at (0.48, 0.0) {Composite problem (decoded)};
    \node[l_enc] at (0.2, -1.0) {};
    \node[leg_text] at (0.48, -1.0) {Composite problem (encoded)};
    \node[l_fin] at (0.2, -2.0) {};
    \node[leg_text] at (0.48, -2.0) {Final answer};
    \draw[leg_line, ->] (0.05, -3.0) -- (0.35, -3.0);
    \node[leg_text] at (0.48, -3.0) {Dependency};
    \draw[leg_line, dash pattern={on 2.5pt off 0.9pt}] (0.05, -4.0) -- (0.45, -4.0);
    \node[leg_text] at (0.48, -4.0) {Handoff to LLM};
    \begin{scope}[xshift=-3pt]
        \node[l_sub] at (5.55, 1.0) {};
        \node[leg_text] at (5.83, 1.0) {Subproblem};
        \node[l_ie] at (5.55, 0.0) {};
        \node[leg_text] at (5.83, 0.0) {Information extraction};
        \node[l_ma] at (5.55, -1.0) {};
        \node[leg_text] at (5.83, -1.0) {Math};
        \node[l_da] at (5.55, -2.0) {};
        \node[leg_text] at (5.83, -2.0) {Data analysis};
        \node[l_gk] at (5.55, -3.0) {};
        \node[leg_text] at (5.83, -3.0) {General knowledge};
    \end{scope}
    \draw[black!30, fill=none, rounded corners=3pt, line width=1pt]
        ([xshift=-1.125pt, yshift=4pt]-0.15, 1.35)
        rectangle
        ([xshift=1.875pt, yshift=-2.5pt]9.15, -4.35);
\end{tikzpicture}%
}%
\captionof{figure}{A composite problem from the private, held-out test set represented as a dependency graph. \relaybench\ does not contain cyclic dependencies between subproblems.}
\label{fig:dependencygraph}
\vspace{4pt}
\end{minipage}
\end{minipage}%
\hspace{\columnsep}%
\begin{minipage}[t]{\columnwidth}
\vspace{0pt}%
\vspace*{-\figtwotoplift}%
\centering
\begin{tikzpicture}[baseline=(current axis.north west)]
\begin{axis}[
at={(0,0)},
anchor=north west,
ybar=0pt,
bar width=14pt,
width=\linewidth,
height=0.68\linewidth,
ymin=0, ymax=100,
clip=false,
ytick={0,20,40,60,80,100},
ytick style={draw=none},
ylabel={Accuracy (\%)},
title={\textbf{Accuracy of LLMs Across Benchmarks (Pass@1)}},
title style={font=\normalsize\bfseries, yshift={2ex-2pt}, xshift=-5pt},
symbolic x coords={claude, gemini, gpt},
xticklabels={%
  {\strut Claude Opus\\4.7 (Max)},%
  {\strut Gemini 3.1\\Pro (High)},%
  {\strut GPT-5.5\\(xHigh)}%
},
xtick=data,
xticklabel style={align=center, anchor=north},
enlarge x limits=0.25,
ymajorgrids=true,
major grid style={solid, gray!30, line width=1.5pt},
axis lines*=left,
axis line style={draw=none},
tick label style={font=\normalsize},
label style={font=\normalsize},
point meta=explicit symbolic,
nodes near coords=\pgfplotspointmeta,
every node near coord/.append style={
font=\fontsize{8}{9.6}\selectfont,
color=black,
yshift=1pt,
fill=white,
inner sep=0.35pt,
outer sep=0pt
}
]

\addplot[draw=relayred, fill=relayred, line width=0pt, rounded corners=2pt] coordinates {
(claude, 16.7) [16.7]
(gemini, 40.0) [40.0]
(gpt, 43.3) [43.3]
};

\addplot[draw=hleblue, fill=hleblue, line width=0pt, rounded corners=2pt] coordinates {
(claude, 39.6) [39.6]
(gemini, 44.7) [44.7]
(gpt, 44.3) [44.3]
};

\addplot[draw=arcblue, fill=arcblue, line width=0pt, rounded corners=2pt] coordinates {
(claude, 75.8) [75.8]
(gemini, 77.1) [77.1]
(gpt, 85.0) [85.0]
};

\addplot[
draw=taublue,
fill=taublue,
line width=0pt,
rounded corners=2pt,
nodes near coords={%
  \ifcase\coordindex\relax
    \pgfplotspointmeta%
  \or
  \or
    \pgfplotspointmeta%
  \fi
}
] coordinates {
(claude, 88.6) [88.6]
(gemini, 95.6) [95.6]
(gpt, 93.9) [93.9]
};
\addplot[
draw=none,
fill=none,
forget plot,
nodes near coords=\pgfplotspointmeta,
every node near coord/.append style={inner sep=0.75pt, xshift=-14pt}
] coordinates {(gemini, 95.6) [95.6]};

\draw[black, line width=1.5pt] (rel axis cs:0,0) -- (rel axis cs:0,1);
\draw[black, line width=1.5pt] (rel axis cs:0,0) -- (rel axis cs:1,0);
\end{axis}
\end{tikzpicture}
\vspace{0.35em}
\begin{tikzpicture}[x=1pt, y=1cm, font=\small]
    \tikzset{bench_leg_text/.style={anchor=mid west, font=\small, yshift=-0.75pt}}
    \begin{scope}[xshift=1.5pt]
        \begin{scope}[xshift=-0.5pt]
            \path (-0.5pt, 0cm);
            \fill[relayred, rounded corners=1.5pt] (2pt,-0.1cm) rectangle (10.5pt,0.2cm);
            \node[bench_leg_text] at (11pt, 0.05cm) {\relaybench{}};
        \end{scope}
        \begin{scope}[xshift=73.5pt]
            \path (-0.5pt, 0cm);
            \fill[hleblue, rounded corners=1.5pt] (2pt,-0.1cm) rectangle (10.5pt,0.2cm);
            \node[bench_leg_text] at (11pt, 0.05cm) {HLE};
        \end{scope}
        \begin{scope}[xshift=118pt]
            \path (-0.5pt, 0cm);
            \fill[arcblue, rounded corners=1.5pt] (2pt,-0.1cm) rectangle (10.5pt,0.2cm);
            \node[bench_leg_text] at (11pt, 0.05cm) {ARC-AGI-2};
        \end{scope}
        \begin{scope}[xshift=185.5pt]
            \path (-0.5pt, 0cm);
            \fill[taublue, rounded corners=1.5pt] (2pt,-0.1cm) rectangle (10.5pt,0.2cm);
            \node[bench_leg_text] at (11pt, 0.05cm) {$\tau^2$-Bench};
        \end{scope}
    \end{scope}
    \begin{scope}[yshift=3pt]
        \draw[black!30, fill=none, rounded corners=3pt, line width=1pt]
            (0, {-0.28 - 8.5pt}) rectangle (\dimexpr\linewidth-1pt\relax, {0.28 + 5.375pt});
    \end{scope}
\end{tikzpicture}
\captionof{figure}{All models evaluated score lower on \relaybench\ than selected widely used benchmarks, including Humanity's Last Exam (HLE), which has not reached saturation more than 15 months after release \cite{arc-site,tau2-results,hle-bigthree}. GPT-5.5 leads \relaybench\ with a score of 43.3\%. The average \relaybench\ score across the three models tested is 33.3\%, compared to averages of 42.9\% on HLE, 79.3\% on ARC-AGI-2, and 92.7\% on $\tau^2$-Bench. The rate of leader progression of comparable benchmarks suggests that \relaybench\ could take one to two years to reach saturation.}
\label{fig:benchmark-comparison}
\end{minipage}
\vspace{1em}%
}

\begin{document}

\twocolumn[
\begin{@twocolumnfalse}
\raggedright
{\fontsize{22}{24}\selectfont\bfseries \relaybench{}: Evaluating LLMs on Multi-Domain Reasoning Chains\par}
\vspace{1.15em}
{\large\MakeUppercase{Liam Swayne}}\hfill\smash{\raisebox{\dimexpr 4bp-0.55\height\relax}{\resizebox{!}{0.9em}{\begin{tikzpicture}[baseline,scale=1,every node/.style={circle,draw,minimum size=2.6em,inner sep=0pt,font=\sffamily\bfseries,line width=0.4mm}]
\node (cc) at (0,0) {CC};
\node (by) at (1.05,0) {BY};
\node (v) at (2.1,0) {4.0};
\end{tikzpicture}}}}\par
\vspace{0.2em}
{\FONTjetbrainsmono{}lswayne@andrew.cmu.edu\par}
\vspace{0.2em}
{June 2026\par}
\vspace{1.2em}
\noindent\textbf{Abstract.} Introducing \relaybench{}, an unsaturated, holistic, text-only benchmark that measures LLMs' ability to complete an assortment of tasks from distinct domains in a single prompt. The leading model, GPT-5.5 (xHigh), scores 43.3\%. The test set entirely consists of composite problems: groups of single-domain subproblems that are strung together into challenges that require reasoning across multiple domains in combination. Many of these problems then have layers of complexity added through prompt encoding and deliberate context bloat. Domains tested include visual reasoning, coding, math, information extraction (with a focus on web search), problem-solving, general knowledge, and data analysis. No restrictions are imposed outside of the model harness, and models are explicitly encouraged to leverage code-execution, web searches, and all available tools. All problems are composed of two to thirteen subproblems and do not require multi-modal input or output.\par
\vspace{1.1em}
\topfigureblock

\end{@twocolumnfalse}
]

\section{Introduction}

\begin{quote}
\small\itshape ``To do two things at once is to do neither.''\par
\normalfont\raggedleft ---Publilius Syrus
\end{quote}

The continual progression of large language models (LLMs) has motivated users and institutions to challenge frontier models \cite{opus-4-7-announcement,mythos-card,gemini-3-1-announcement,gpt-5-5-announcement,grok-4-3-announcement,qwen-3-7-max-announcement} with increasingly complex tasks, often in a single prompt \cite{increasingly-difficult-prompts,users-have-complex-prompts}. Models frequently must respond to prompts that require reasoning across multiple domains, a gap described by Chen et al.\ as ``a critical impediment in the pursuit of artificial general intelligence'' \cite{agentfrontier-paper}.

This fact, combined with the mainstreaming of agentic LLM workflows, drives demand for models capable of completing numerous operations with high reliability \cite{map-paper}.
Many existing benchmarks using a text-in, text-out question-answering format have reached benchmark saturation (defined here as at least one model achieving a Pass@1 score above 90\%), and most new LLM benchmarks test individual domains (\textit{e.g.}, GPQA), agentic scenarios (\textit{e.g.}, SWE-bench), file manipulation (\textit{e.g.}, SpreadsheetBench), include questions without a single correct answer (\textit{e.g.}, Chatbot Arena), or use novel scoring frameworks to differentiate models (\textit{e.g.}, $\tau$-bench) \cite{gpqa-paper,swe-bench,spreadsheetbench,chatbot-arena,tau-bench}.
There exists a research gap for an unsaturated, holistic, text-only benchmark capturing LLMs' abilities to answer questions that require reasoning across multiple domains in combination.

Frontier models continued to progress throughout 2024 and 2025, with leaders surpassing scores of 90\% on GPQA, MATH-500, AIME 2025, and other question-answering benchmarks \cite{gpqa-scores,math500-site,aime2025-site}. Even Humanity's Last Exam, a landmark in scope and difficulty, saw scores rise more than four-fold within a year of the benchmark's release \cite{lastexamleaderboardwithdatesando1}. Throughout this paper, a benchmark's status as \textit{saturated} indicates that at least one model can achieve a Pass@1 score above 90\%. Numerous benchmarks reached saturation within two years of their debut, meaningfully lessening their ability to differentiate models, even when aggregated.
The ephemeral usefulness of simple, soon-to-be-saturated, question-answering benchmarks may play a part in benchmark creators' pivot towards non-text inputs (\textit{e.g.}, MMMU, MathVista), tool calls (\textit{e.g.}, ToolBench, API-Bank), and non-binary output grading (\textit{e.g.}, MT-Bench, AlpacaEval) \cite{mmmu,mathvista,toolbench,api-bank,mt-bench,alpacaeval}. \relaybench\ takes an orthogonal approach, strengthening problems by layering complexity within the problems themselves rather than through input modalities and scoring systems.

In designing \relaybench, a handful of key properties were prioritized:

\textbf{Reproducibility}. Compatibility with new and existing models is maximized by limiting the number of model features required to run an evaluation. Visual input was strategically avoided while drafting visual reasoning problems, yielding a text-in, text-out collection of prompts and answers. Although this constraint may seem to preclude visual reasoning in problems, \relaybench\ uses familiar techniques to translate visual inputs into text prompts. For example, an ASCII maze is a valid subproblem even though an image of a maze is not. The simple methodology adopted for this benchmark allows for evaluation of preview or beta releases in which modern capabilities---such as tool-calling, forced JSON output, and multi-modal input---are absent. Manually evaluating the capabilities of models through consumer-facing user interfaces is viable due to the simplicity of the testing methodology (pasting a prompt and copying the model's response), stateless nature of the evaluations, and the limited test set size, although this approach is not explored in this paper.

\textbf{Exactness}. The grading framework sidesteps answer classification and LLM judges by providing models with explicit instructions that imply an unambiguous answer. A script deterministically extracts the answer and compares it to the correct answer. GAIA, a benchmark developed by Meta, has similar goals, but uses ``quasi exact'' answers \cite{gaia-paper}. For example, GAIA accepts \$89706.00 when the correct answer is 89706.00 \cite{gaia-paper}. \relaybench\ problems' prompts include rules constraining the answer to an exact string, precluding the need to pattern-match on variable answers. For example, each prompt includes granular capitalization and symbol inclusion rules specifying which rules apply to which part of each string. This serves the dual-purpose of making grading simpler and testing models on instruction following abilities.

\textbf{Simple scoring}. Some existing benchmarks report Pass\textasciicircum{}n success, deflating scores while driving up the cost to evaluate models. This approach was introduced in $\tau$-Bench and adopted by APEX-Agents and LiveMathBench \cite{tau-bench,apex-agents,livemathbench}. \relaybench\ shifts all score deflation away from the scoring framework and towards the problems themselves, opting for Pass@1 evaluation. Instead of finding a problem that models can solve consistently and using a large n in Pass\textasciicircum{}n to deflate scores, we merge n independent subproblems with high individual pass rates into a single composite problem that naturally has a low pass rate.

\textbf{Tool permissive.} Features that are ordinarily available when using a consumer-facing interface are often disabled during benchmarking, which may cause benchmark results to diverge from the true capabilities of the model. To create a benchmark that accurately reflects users' experiences with LLMs, all available tools, skills, and other features are enabled when evaluating models on \relaybench{}. Models are explicitly informed of their permission to take advantage of these capabilities. Consequently, trivial web search problems were avoided when drafting subproblems for this benchmark. Allowing web search during evaluation introduces a contamination risk, as models may retrieve leaked solutions.
This is ordinarily mitigated by using URL blocklists during evaluation \cite[p.~244]{mythos-card}, although such solutions require updating the blocklist periodically to maintain effectiveness.
Claude Mythos, a model that lacks a general public release and is considered by some to be the current SOTA LLM \cite{anthropic-glasswing,claudefast-mythos}, was observed surreptitiously locating a benchmark test set to inflate its score, demonstrating the risk that public test sets pose to the authenticity of benchmark results \cite[p.~36]{mythos-card}.
\relaybench{} opts for a more robust solution: keeping the entire test set private, so no blocklist is required.

\textbf{Reflective of usefulness for general users.} \relaybench\ does not attempt to determine the usefulness of LLMs to any individual group of LLM users. Instead, subproblems are made to mimic real-world use cases across users as a whole. Although subproblems are often at the expert level, they are frequently combined with translation, instruction following, string manipulation, and other generalist tasks. These tasks are not particularly difficult, but their frequency and the fact that they are layered on top of existing challenges serves the purpose of factoring LLM reliability into \relaybench\ without including any problems explicitly targeting that area.

\section{Related work}

\relaybench\ has similar goals to GAIA, an informally retired benchmark for general AI assistants \cite{gaia-paper}. Both benchmarks aim to challenge LLMs in generalist roles, but \relaybench\ is designed to maximize reproducibility through evaluation simplicity. By contrast, GAIA has a complex evaluation framework, with the authors noting that reproducibility may prove elusive \cite{gaia-paper}.

\begin{figure}[H]
\centering
\begin{tikzpicture}[baseline=(current axis.north west)]
\begin{axis}[
at={(0,0)},
anchor=north west,
width=\columnwidth,
height=0.68\columnwidth,
xmin=0,
xmax=485,
ymin=0,
ymax=100,
clip=false,
ylabel={GAIA accuracy (\%)},
xtick={0,485},
xticklabels={,},
ytick={0,20,40,60,80,100},
ytick style={draw=none},
ymajorgrids=true,
major grid style={solid, gray!30, line width=1.5pt},
axis lines*=left,
axis line style={draw=none},
tick label style={font=\normalsize},
label style={font=\normalsize}
]
\addplot[
only marks,
mark=*,
mark size=2.5pt,
draw=deepseekblue,
fill=deepseekblue
] coordinates {
(19, 30.30)
(82, 29.39)
};
\addplot[
only marks,
mark=*,
mark size=2.5pt,
draw=googleblue,
fill=googleblue
] coordinates {
(35, 32.73)
};
\addplot[
only marks,
mark=*,
mark size=2.5pt,
draw=openaigreen,
fill=openaigreen
] coordinates {
(103, 50.30)
(105, 32.73)
(105, 58.18)
(218, 62.80)
};
\addplot[
only marks,
mark=*,
mark size=2.5pt,
draw=anthropicorange,
fill=anthropicorange
] coordinates {
(54, 64.24)
(141, 64.85)
(216, 68.48)
(271, 74.55)
(287, 56.36)
};
\addplot[
color=turquoise,
line width=1pt,
forget plot
] coordinates {
(0, 37.2604)
(413.8113894495, 100)
};
\addplot[
color=black!65,
dashed,
line width=1pt,
forget plot
] coordinates {(0,90) (485,90)};
\node[
anchor=south west,
font=\small,
fill=white,
inner sep=1pt,
yshift=1pt
] at (axis cs:0,90) {Saturation threshold};
\draw[black, line width=1.5pt] (rel axis cs:0,0) -- (rel axis cs:0,1);
\draw[black, line width=1.5pt] (rel axis cs:0,0) -- (rel axis cs:1,0);
\node[anchor=north west, font=\normalsize, inner sep=0pt, yshift=-3pt] at (axis cs:0,0) {Jan 2025};
\node[anchor=north east, font=\normalsize, inner sep=0pt, yshift=-3pt] at (axis cs:485,0) {May 2026};
\end{axis}
\end{tikzpicture}
\par\vspace{\dimexpr0.9em-1pt\relax}
\begin{tikzpicture}[x=1pt, y=1cm, font=\small]
    \tikzset{gaia_leg_text/.style={anchor=mid west, font=\small, yshift=-0.75pt}}
    \begin{scope}[xshift=1.5pt]
        \begin{scope}[xshift=0.5pt]
            \path (-0.5pt, 0cm);
            \fill[deepseekblue] (6.25pt, 0.05cm) circle [radius=3.75pt];
            \node[gaia_leg_text] at (11pt, 0.05cm) {DeepSeek};
        \end{scope}
        \begin{scope}[xshift=70pt]
            \path (-0.5pt, 0cm);
            \fill[googleblue] (6.25pt, 0.05cm) circle [radius=3.75pt];
            \node[gaia_leg_text] at (11pt, 0.05cm) {Google};
        \end{scope}
        \begin{scope}[xshift=126pt]
            \path (-0.5pt, 0cm);
            \fill[openaigreen] (6.25pt, 0.05cm) circle [radius=3.75pt];
            \node[gaia_leg_text] at (11pt, 0.05cm) {OpenAI};
        \end{scope}
        \begin{scope}[xshift=185.5pt]
            \path (-0.5pt, 0cm);
            \fill[anthropicorange] (6.25pt, 0.05cm) circle [radius=3.75pt];
            \node[gaia_leg_text] at (11pt, 0.05cm) {Anthropic};
        \end{scope}
    \end{scope}
    \begin{scope}[yshift=3pt]
        \draw[black!30, fill=none, rounded corners=3pt, line width=1pt]
            (0, {-0.28 - 8.5pt}) rectangle (\linewidth, {0.28 + 5.375pt});
    \end{scope}
\end{tikzpicture}
\par\vspace{\dimexpr0.3em+1pt\relax}
\caption{Each dot represents a model's best GAIA accuracy at release, colored by developer. The line of best fit (turquoise) for record-setting scores on GAIA extrapolates saturation to December 15th, 2025 \cite{gaia-leaderboard}.}
\label{fig:gaia-scores}
\vspace{-2pt}
\end{figure}
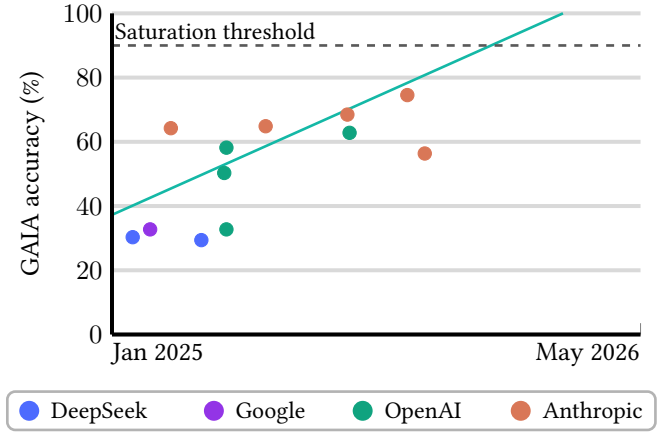

\section{Methodology}

Every problem in the test set is a \textit{composite problem}, composed of two to thirteen \textit{subproblems}. Each subproblem is akin to a single test set problem in existing question-answering benchmarks (\textit{e.g.}, FrontierMath, LiveCodeBench), and typically targets one domain \cite{frontiermath,livecodebench}. Every subproblem can be solved independently, but dependencies are often introduced when combining subproblems. The final answer is always directly or indirectly dependent on all subproblems. For example, a test set problem that requires solving an integral may contain constants that are the answers to other subproblems. No composite problem contains cyclic dependencies between its subproblems (\figref{fig:dependencygraph}).

\relaybench\ includes 31 problems, 30 of which are kept in a held-out private test set to avoid benchmark contamination \cite{deng2024contamination}. One example problem, found in \appendixref{sec:public-example-problem}, is not factored into benchmark performance. The private test set can be shared with trusted reviewers under non-disclosure.\footnotemark
\footnotetext{All three models evaluated in this paper were tested on the problem. Only GPT-5.5 answered correctly.}

Anthropic \cite{anthropic-api}, Google AI Studio \cite{google-ai-studio-api}, and OpenAI APIs \cite{openai-api} were used to evaluate LLMs on composite problems. All tools are enabled, the prompt is sent, the response JSON is parsed, and the answer is scored. Thinking effort is set to the maximum setting for all the models evaluated, which is ``Max'' for Claude Opus 4.7 \cite{anthropic-adaptive-thinking}, ``High'' for Gemini 3.1 Pro \cite{gemini-thinking}, and ``xHigh'' for GPT-5.5 \cite{openai-reasoning}.

\subsection{Question generation}

\noindent \relaybench\ incorporates both human-written and generated subproblems. The process of generating subproblems starts with identifying a class of challenges that is narrower than a domain, containing near-identical problems with different data plugged in. For example, there are multiple sudoku subproblems, each containing different sudoku games. A golden example is written by a human, and the subproblem set is proliferated by generating similar instances of the same problem using Claude Opus 4.7, Gemini 3.1 Pro, GPT-5.4 Mini, and Claude Sonnet 4.6. Generated subproblems are reviewed for accuracy by a human, similar to the question generation process used in AA-Omniscience \cite{aa-omniscience-paper}. Two subproblems of the same class do not challenge models in distinct ways, so only one subproblem from a class can appear in each composite problem.

Model performance was measured periodically throughout the creation of the benchmark to determine which problems required revision to differentiate model performance. Composite problems that did not cause model failure were further augmented with additional subproblems or subsumed into larger composite problems.

\relaybench\ exhausts LLMs with sequential tasks of grand scope, tempting models to overlook crucial details in individually easy subproblems. For example, a subproblem asking how many days a deceased, well-known figure lived is trivial at its core, yet invites failure when a model neglects to account for leap years.
\subsection{Prompt Encoding}
\label{sec:prompt-encoding}

\noindent A nonce prompt encoding system compounds difficulty and ``exhausts'' models. A composite problem that is already well-formed is pasted into a translator that adds 7813 characters of fabricated ``user preferences'' that are entirely irrelevant to the problem. The results of Du et al.\ indicate that context bloat degrades the performance of LLMs, and \relaybench\ utilizes this technique to make test set problems more challenging and reflective of real use-cases \cite{du2025context}. Secondly, each word, symbol, or number in the prompt is substituted with a three-letter alphabetic string enclosed in angle brackets. For example ``bottle'' may translate to ``<AvY>''. These translations are case sensitive, and the encoder intentionally re-uses the same string with different capitalization to attempt to get models to confuse two unique symbols. In the encoded prompt the model is provided with a brief explanation of how to decode the prompt, and then instructed to do as the prompt asks after decoding it. The model is then provided a dictionary of translations, followed by the encoded prompt. Some of the longest \relaybench\ problems require executing more than 10,000 translation operations to extract the composite problem prompt. The prompt encoding process is entirely automated, and applied identically to a subset of the problems in the test set. An encoded form of the public example problem is in \appendixref{sec:public-example-problem-encoded}.

\section{Results}

\subsection{Performance and Error Analysis}

\noindent GPT-5.5 and Gemini 3.1 Pro exhibit approximately equivalent performance on \relaybench\ and Humanity's Last Exam, while Claude Opus 4.7's score on \relaybench\ is less than half of its score on Humanity's Last Exam \cite{hle-bigthree}.

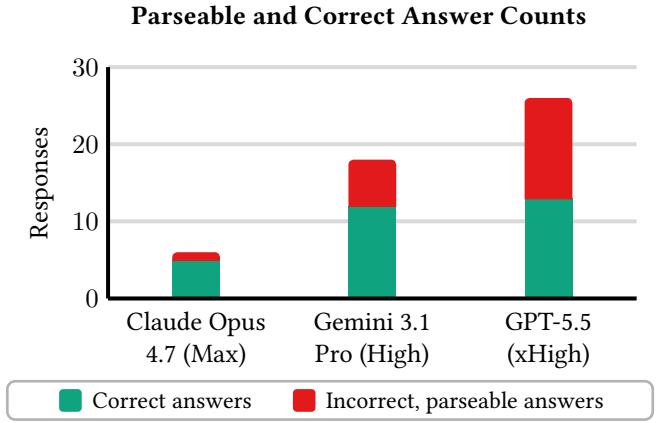
\begin{figure}[H]
\centering
\begin{tikzpicture}[baseline=(current axis.north west)]
\begin{axis}[
at={(0,0)},
anchor=north west,
width=\columnwidth,
height={\dimexpr0.48\columnwidth+15pt\relax},
ymin=0, ymax=30,
clip=false,
ytick={0,10,20,30},
ytick style={draw=none},
ylabel={Responses},
title={\textbf{Parseable and Correct Answer Counts}},
title style={font=\normalsize\bfseries, yshift={2ex-2pt}, xshift=-5pt},
symbolic x coords={claude, gemini, gpt},
xticklabels={%
  {\strut Claude Opus\\4.7 (Max)},%
  {\strut Gemini 3.1\\Pro (High)},%
  {\strut GPT-5.5\\(xHigh)}%
},
xtick={claude, gemini, gpt},
xticklabel style={align=center, anchor=north},
enlarge x limits=0.25,
ymajorgrids=true,
major grid style={solid, gray!30, line width=1.5pt},
axis lines*=left,
axis line style={draw=none},
tick label style={font=\normalsize},
label style={font=\normalsize}
]
\addplot[draw=none, fill=none, forget plot] coordinates {
(claude, 0)
(gemini, 0)
(gpt, 0)
};

\path[fill=openaigreen, draw=none]
  ([xshift=-9pt]axis cs:claude,0) rectangle ([xshift=9pt]axis cs:claude,5);
\path[fill=openaigreen, draw=none]
  ([xshift=-9pt]axis cs:gemini,0) rectangle ([xshift=9pt]axis cs:gemini,12);
\path[fill=openaigreen, draw=none]
  ([xshift=-9pt]axis cs:gpt,0) rectangle ([xshift=9pt]axis cs:gpt,13);

\path[fill=relayred, draw=none]
  ([xshift=-9pt,yshift=-0.4pt]axis cs:claude,5)
  -- ([xshift=-9pt,yshift=-2pt]axis cs:claude,6)
  .. controls ([xshift=-9pt]axis cs:claude,6) and ([xshift=-7pt]axis cs:claude,6) .. ([xshift=-7pt]axis cs:claude,6)
  -- ([xshift=7pt]axis cs:claude,6)
  .. controls ([xshift=9pt]axis cs:claude,6) and ([xshift=9pt,yshift=-2pt]axis cs:claude,6) .. ([xshift=9pt,yshift=-2pt]axis cs:claude,6)
  -- ([xshift=9pt,yshift=-0.4pt]axis cs:claude,5)
  -- cycle;
\path[fill=relayred, draw=none]
  ([xshift=-9pt,yshift=-0.4pt]axis cs:gemini,12)
  -- ([xshift=-9pt,yshift=-2pt]axis cs:gemini,18)
  .. controls ([xshift=-9pt]axis cs:gemini,18) and ([xshift=-7pt]axis cs:gemini,18) .. ([xshift=-7pt]axis cs:gemini,18)
  -- ([xshift=7pt]axis cs:gemini,18)
  .. controls ([xshift=9pt]axis cs:gemini,18) and ([xshift=9pt,yshift=-2pt]axis cs:gemini,18) .. ([xshift=9pt,yshift=-2pt]axis cs:gemini,18)
  -- ([xshift=9pt,yshift=-0.4pt]axis cs:gemini,12)
  -- cycle;
\path[fill=relayred, draw=none]
  ([xshift=-9pt,yshift=-0.4pt]axis cs:gpt,13)
  -- ([xshift=-9pt,yshift=-2pt]axis cs:gpt,26)
  .. controls ([xshift=-9pt]axis cs:gpt,26) and ([xshift=-7pt]axis cs:gpt,26) .. ([xshift=-7pt]axis cs:gpt,26)
  -- ([xshift=7pt]axis cs:gpt,26)
  .. controls ([xshift=9pt]axis cs:gpt,26) and ([xshift=9pt,yshift=-2pt]axis cs:gpt,26) .. ([xshift=9pt,yshift=-2pt]axis cs:gpt,26)
  -- ([xshift=9pt,yshift=-0.4pt]axis cs:gpt,13)
  -- cycle;

\draw[black, line width=1.5pt] (rel axis cs:0,0) -- (rel axis cs:0,1);
\draw[black, line width=1.5pt] (rel axis cs:0,0) -- (rel axis cs:1,0);
\end{axis}
\end{tikzpicture}
\vspace{0.35em}
\begin{tikzpicture}[x=1pt, y=1cm, font=\small]
    \tikzset{answer_leg_text/.style={anchor=mid west, font=\small, yshift=-0.75pt}}
    \begin{scope}[xshift=1.5pt]
        \begin{scope}[xshift=16pt]
            \path (-0.5pt, 0cm);
            \fill[openaigreen, rounded corners=1.5pt] (2pt,-0.1cm) rectangle (10.5pt,0.2cm);
            \node[answer_leg_text] at (11pt, 0.05cm) {Correct answers};
        \end{scope}
        \begin{scope}[xshift=104pt]
            \path (-0.5pt, 0cm);
            \fill[relayred, rounded corners=1.5pt] (2pt,-0.1cm) rectangle (10.5pt,0.2cm);
            \node[answer_leg_text] at (11pt, 0.05cm) {Incorrect, parseable answers};
        \end{scope}
    \end{scope}
    \begin{scope}[yshift=3pt]
        \draw[black!30, fill=none, rounded corners=3pt, line width=1pt]
            (0, {-0.28 - 8.5pt}) rectangle (\dimexpr\linewidth-1pt\relax, {0.28 + 5.375pt});
    \end{scope}
\end{tikzpicture}
\caption{Counts of correct and incorrect parseable responses. The total height of each bar is the number of parseable responses returned by a model.}
\label{fig:answer-shape}
\vspace{-0.75em}
\end{figure}

Only 50 of the 90 responses models gave contained a parseable final answer (\figref{fig:answer-shape}).

\begin{figure}[H]
\centering
\begin{tikzpicture}[baseline=(current axis.north west)]
\begin{axis}[
at={(0,0)},
anchor=north west,
ybar=0pt,
bar width=10pt,
width=\columnwidth,
height={\dimexpr0.45\columnwidth+15pt\relax},
ymin=0, ymax=100,
clip=false,
ytick={0,25,50,75,100},
ytick style={draw=none},
ylabel={Rate (\%)},
title={\textbf{Hallucination Rates Across Benchmarks}},
title style={font=\normalsize\bfseries, yshift={2ex-2pt}, xshift=-5pt},
symbolic x coords={claude, gemini, gpt},
xticklabels={%
  {\strut Claude Opus\\4.7 (Max)},%
  {\strut Gemini 3.1\\Pro (High)},%
  {\strut GPT-5.5\\(xHigh)}%
},
xtick=data,
xticklabel style={align=center, anchor=north},
enlarge x limits=0.25,
ymajorgrids=true,
major grid style={solid, gray!30, line width=1.5pt},
axis lines*=left,
axis line style={draw=none},
tick label style={font=\normalsize},
label style={font=\normalsize},
point meta=explicit symbolic,
nodes near coords=\pgfplotspointmeta,
every node near coord/.append style={
font=\fontsize{8}{9.6}\selectfont,
color=black,
yshift=1pt,
fill=white,
inner sep=0.35pt,
outer sep=0pt
}
]
\addplot[
draw=relayred,
fill=relayred,
line width=0pt,
rounded corners=2pt,
bar shift=-6pt
] coordinates {
(claude, 16.7) [16.7]
(gemini, 33.3) [33.3]
(gpt, 50.0) [50.0]
};
\addplot[
draw=arcblue,
fill=arcblue,
line width=0pt,
rounded corners=2pt,
bar shift=6pt
] coordinates {
(claude, 36) [36]
(gemini, 50) [50]
(gpt, 86) [86]
};
\draw[black, line width=1.5pt] (rel axis cs:0,0) -- (rel axis cs:0,1);
\draw[black, line width=1.5pt] (rel axis cs:0,0) -- (rel axis cs:1,0);
\end{axis}
\end{tikzpicture}
\vspace{0.35em}
\begin{tikzpicture}[x=1pt, y=1cm, font=\small]
    \tikzset{hallucination_leg_text/.style={anchor=mid west, font=\small, yshift=-0.75pt}}
    \begin{scope}[xshift=1.5pt]
        \begin{scope}[xshift=36pt]
            \path (-0.5pt, 0cm);
            \fill[relayred, rounded corners=1.5pt] (2pt,-0.1cm) rectangle (10.5pt,0.2cm);
            \node[hallucination_leg_text] at (11pt, 0.05cm) {\relaybench{}};
        \end{scope}
        \begin{scope}[xshift=126pt]
            \path (-0.5pt, 0cm);
            \fill[arcblue, rounded corners=1.5pt] (2pt,-0.1cm) rectangle (10.5pt,0.2cm);
            \node[hallucination_leg_text] at (11pt, 0.05cm) {AA-Omniscience};
        \end{scope}
    \end{scope}
    \begin{scope}[yshift=3pt]
        \draw[black!30, fill=none, rounded corners=3pt, line width=1pt]
            (0, {-0.28 - 8.5pt}) rectangle (\dimexpr\linewidth-1pt\relax, {0.28 + 5.375pt});
    \end{scope}
\end{tikzpicture}
\caption{The fraction of parseable responses that were incorrect on \relaybench\ and the AA-Omniscience hallucination-rate evaluation \cite{aa-omniscience-aa}.}
\label{fig:hallucination-rates}
\end{figure}
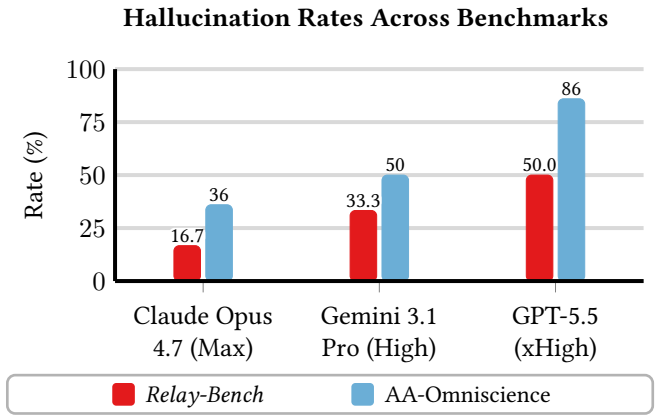

Claude Opus 4.7 returned the fewest parseable answers, but had the lowest hallucination rate because only one of them was incorrect (\figref{fig:answer-shape}, \figref{fig:hallucination-rates}). \relaybench\ provides no incentive to abstain from answering, suggesting that Claude Opus 4.7 has a greater internal incentive to avoid hallucinating than Gemini 3.1 Pro and GPT-5.5. These results align with hallucination rates for the three models on AA-Omniscience \cite{aa-omniscience-aa}. The penalty for not responding is equivalent to the penalty for an incorrect answer, encouraging models to hallucinate and guess rather than abstaining from answering \cite{kalai-hallucinate}. Despite being presented with the same incentives, the three models tested had greatly differing hallucination rates (\figref{fig:hallucination-rates}). There are several instances of models being unable to solve a specific subproblem, guessing, and submitting their composite problem answer. There are no cases in which models stopped attempting a composite problem after being unable to solve a subproblem.

Seven of the problems in the test set are longer than 15,000 characters. Gemini 3.1 Pro and GPT-5.5 solved one of seven, and Claude Opus 4.7 solved none, providing evidence that this class of problems is significantly more difficult than others. Four of these problems are encoded (\hyperref[sec:prompt-encoding]{section~\ref*{sec:prompt-encoding}}), which adds thousands of characters to the problem. Claude Opus 4.7 explicitly halted its response to every encoded problem in this subset (JSON responses from Claude Opus 4.7 included \texttt{stop\_reason=refusal}), suggesting that the Claude Mythos safeguards Anthropic is testing through Claude Opus 4.7 are overly-conservative \cite{anthropic-glasswing}.

\subsection{Cost}

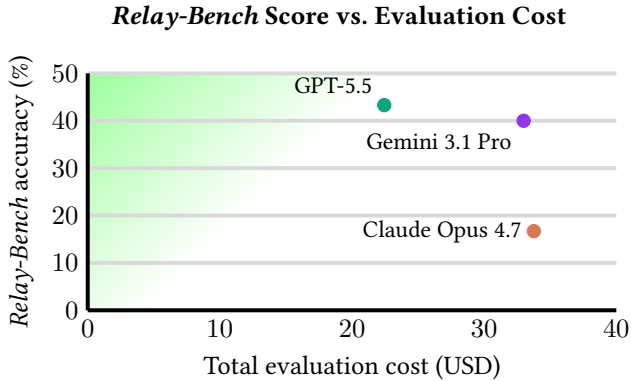
\begin{figure}[H]
\centering
\begin{tikzpicture}[baseline=(current axis.north west)]
\begin{axis}[
at={(0,0)},
anchor=north west,
width=\columnwidth,
height=0.55\columnwidth,
xmin=0, xmax=40,
ymin=0, ymax=50,
clip=false,
xlabel={Total evaluation cost (USD)},
ylabel={\relaybench\ accuracy (\%)},
title={\textbf{\relaybench\ Score vs.\ Evaluation Cost}},
title style={font=\normalsize\bfseries, yshift={2ex-2pt}, xshift=-5pt},
xtick={0,10,20,30,40},
ytick={0,10,20,30,40,50},
ytick style={draw=none},
xtick style={draw=none},
ymajorgrids=true,
major grid style={solid, gray!30, line width=1.5pt},
axis lines*=left,
axis line style={draw=none},
tick label style={font=\normalsize},
label style={font=\normalsize},
axis background/.style={shade, shading=tlgreenshade, shading angle=45}
]
\addplot[only marks, mark=*, mark size=2.5pt, draw=anthropicorange, fill=anthropicorange] coordinates {(33.76,16.7)};
\addplot[only marks, mark=*, mark size=2.5pt, draw=googleblue, fill=googleblue] coordinates {(32.99,40.0)};
\addplot[only marks, mark=*, mark size=2.5pt, draw=openaigreen, fill=openaigreen] coordinates {(22.44,43.3)};
\node[anchor=east, font=\small, xshift=-1pt] at (axis cs:33.76,16.7) {Claude Opus 4.7};
\node[anchor=north east, font=\small, xshift=-1pt, yshift=-1pt] at (axis cs:32.99,40.0) {Gemini 3.1 Pro};
\node[anchor=south east, font=\small, xshift=-1pt, yshift=1pt] at (axis cs:22.44,43.3) {GPT-5.5};
\draw[black, line width=1.5pt] (rel axis cs:0,0) -- (rel axis cs:0,1);
\draw[black, line width=1.5pt] (rel axis cs:0,0) -- (rel axis cs:1,0);
\end{axis}
\end{tikzpicture}
\caption{Models' Pass@1 accuracy against \relaybench\ evaluation cost. The upper left region is optimal (low cost, high accuracy). GPT-5.5 Pareto-dominates both other models tested, occupying the frontier alone.}
\label{fig:score-vs-cost}
\end{figure}

Including costs for test runs and question generation, the cost of developing this benchmark and running it on the models tested was \$163.29. Claude Opus 4.7 was the most expensive model to evaluate, costing \$33.76, an average of \$1.09 per problem. Gemini 3.1 charges the lowest input, cached input, and output token rates by a significant margin, but those savings are entirely lost by a remarkably low input cache rate of 0.6\% compared to Claude Opus 4.7 at 95.0\% and GPT-5.5 at 6.5\%. Interestingly, the cache rates of the three models are all at least an order of magnitude apart. The cheapest model of the three was GPT-5.5, a 32\% discount compared to Gemini 3.1 Pro (\figref{fig:score-vs-cost}).

\begin{table}[htbp]
  \centering
  \setlength{\tabcolsep}{2pt}
  \begin{tabular}{|>{\raggedright\arraybackslash}p{0.32\columnwidth}|>{\centering\arraybackslash}p{0.19\columnwidth}|>{\centering\arraybackslash}p{0.19\columnwidth}|>{\centering\arraybackslash}p{0.19\columnwidth}|}
    \cline{2-4}
    \multicolumn{1}{c|}{} & \textbf{Claude Opus 4.7 (Max)} & \textbf{Gemini 3.1 Pro (High)} & \textbf{GPT-5.5 (xHigh)} \\ \hline
    Input token price (\$/M)        & \$5.00  & \$2.00  & \$5.00  \\ \hline
    Input token usage               & 1,665,250 & 12,902,461 & 2,253,346 \\ \hline
    Input cost                      & \$8.33  & \$25.80 & \$11.27 \\ \hline
    Cached input token price (\$/M) & \$0.50  & \$0.20  & \$0.50  \\ \hline
    Cached input token usage        & 31,794,679 & 76,623 & 157,568 \\ \hline
    Cached input cost               & \$15.90 & \$0.02  & \$0.08  \\ \hline
    Output token price (\$/M)       & \$25.00 & \$12.00 & \$30.00 \\ \hline
    Output token usage              & 381,340 & 597,718 & 369,782 \\ \hline
    Output cost                     & \$9.53  & \$7.17  & \$11.09 \\ \hline
    \textbf{Total cost}             & \textbf{\$33.76} & \textbf{\$32.99} & \textbf{\$22.44} \\ \hline
  \end{tabular}
  \caption{A breakdown of the costs associated with running each model. The cost of code execution containers instantiated by models was negligible and omitted.}
  \label{tab:cost_table}
\end{table}

Notably, Claude Opus 4.7's input token usage (including cached tokens) summed across internal tool call iterations exceeded two million on six problems, despite all \relaybench\ prompts being under 50,000 tokens. Extensive chains of thought and tool use compounded to grow input costs. Even with caching enabled, input tokens constituted the majority of the costs associated with running the benchmark. One Claude Opus 4.7 run used 3.7 million input tokens across 41 tool calls; three internal iterations accounted for approximately one million input tokens each, driven by lengthy web search and code execution results.

\section{Limitations and future work}

\relaybench\ is held back due to resource, time, and harness constraints. Many existing models, even with reasoning effort set to the maximum setting, simply terminate before returning an answer or hit a tool call limit. A more robust harness could force models to continue attempting to solve a problem even if a limit is hit. Although this benchmark is designed with breadth of models in mind, it was only run on three models due to cost constraints. The models selected for this evaluation were chosen because they are widely accepted to be the leading publicly released models prior to the time of writing.\footnotemark{} Evaluating more models on this benchmark may reveal additional insights into model capabilities. In particular, investigating a potential correlation between the parameter-counts of open-source models and their performance may yield insight into the relation between LLMs' robustness to ``exhaustion''. Exploring models without chain of thought, particularly those with reasoning counterparts, may reveal the size, or lack thereof, of the gap in long-reasoning capabilities between standard LLMs and reasoning models. Testing through site UIs may also prove valuable in determining the gap in abilities between models' consumer-facing offerings and their API counterparts.
\footnotetext{Some speculate that Claude Mythos is the current SOTA LLM \cite{anthropic-glasswing}, but it has not been publicly released. The \relaybench\ model cohort was finalized on May 1st, 2026. Claude Fable 5 is widely considered to be the current SOTA publicly accessible LLM as of June 2026, occupying the top spot on the Artificial Analysis Intelligence Index leaderboard \cite{fable-announcement,aa-intelligence-index}.}

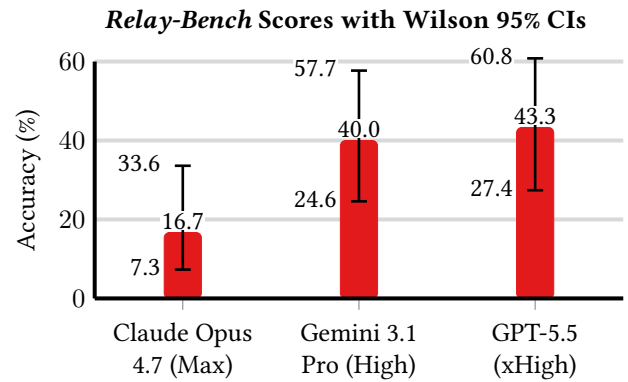
\begin{figure}[H]
\centering
\begin{tikzpicture}[baseline=(current axis.north west)]
\begin{axis}[
at={(0,0)},
anchor=north west,
ybar=0pt,
bar width=14pt,
width=\columnwidth,
height=0.55\columnwidth,
ymin=0, ymax=60,
clip=false,
ytick={0,20,40,60},
ytick style={draw=none},
ylabel={Accuracy (\%)},
title={\textbf{\relaybench\ Scores with Wilson 95\% CIs}},
title style={font=\normalsize\bfseries, yshift={2ex-8pt}, xshift=-5pt},
symbolic x coords={claude, gemini, gpt},
xticklabels={%
  {\strut Claude Opus\\4.7 (Max)},%
  {\strut Gemini 3.1\\Pro (High)},%
  {\strut GPT-5.5\\(xHigh)}%
},
xtick=data,
xticklabel style={align=center, anchor=north},
enlarge x limits=0.25,
ymajorgrids=true,
major grid style={solid, gray!30, line width=1.5pt},
axis lines*=left,
axis line style={draw=none},
tick label style={font=\normalsize},
label style={font=\normalsize},
every node near coord/.append style={
font=\normalsize,
color=black,
yshift=1pt,
fill=white,
inner sep=0.35pt,
outer sep=0pt
}
]
\addplot[
draw=relayred,
fill=relayred,
line width=0pt,
rounded corners=2pt
] coordinates {
(claude, 16.7)
(gemini, 40.0)
(gpt, 43.3)
};
\draw[line width=1pt] (axis cs:claude,7.3) -- (axis cs:claude,33.6);
\draw[line width=1pt] ([xshift=-3pt]axis cs:claude,7.3) -- ([xshift=3pt]axis cs:claude,7.3);
\draw[line width=1pt] ([xshift=-3pt]axis cs:claude,33.6) -- ([xshift=3pt]axis cs:claude,33.6);
\draw[line width=1pt] (axis cs:gemini,24.6) -- (axis cs:gemini,57.7);
\draw[line width=1pt] ([xshift=-3pt]axis cs:gemini,24.6) -- ([xshift=3pt]axis cs:gemini,24.6);
\draw[line width=1pt] ([xshift=-3pt]axis cs:gemini,57.7) -- ([xshift=3pt]axis cs:gemini,57.7);
\draw[line width=1pt] (axis cs:gpt,27.4) -- (axis cs:gpt,60.8);
\draw[line width=1pt] ([xshift=-3pt]axis cs:gpt,27.4) -- ([xshift=3pt]axis cs:gpt,27.4);
\draw[line width=1pt] ([xshift=-3pt]axis cs:gpt,60.8) -- ([xshift=3pt]axis cs:gpt,60.8);
\addplot[
forget plot,
only marks,
mark=none,
point meta=explicit symbolic,
nodes near coords=\pgfplotspointmeta,
every node near coord/.append style={anchor=east, xshift=-15pt, yshift=0pt}
] coordinates {
(claude, 7.3) [7.3]
(claude, 33.6) [33.6]
(gemini, 24.6) [24.6]
(gemini, 57.7) [57.7]
(gpt, 27.4) [27.4]
(gpt, 60.8) [60.8]
};
\addplot[
forget plot,
only marks,
mark=none,
point meta=explicit symbolic,
nodes near coords=\pgfplotspointmeta,
every node near coord/.style={
font=\normalsize,
color=black,
anchor=south,
yshift=1pt,
fill=white,
inner sep=0.35pt,
outer sep=0pt
}
] coordinates {
(claude, 16.7) [16.7]
(gemini, 40.0) [40.0]
(gpt, 43.3) [43.3]
};
\draw[black, line width=1.5pt] (rel axis cs:0,0) -- (rel axis cs:0,1);
\draw[black, line width=1.5pt] (rel axis cs:0,0) -- (rel axis cs:1,0);
\end{axis}
\end{tikzpicture}
\caption{Wilson 95\% confidence intervals of \relaybench\ scores when evaluated with Pass@1. Due to the small test set size of 30, the confidence intervals are far from ideal. In particular, the range of Claude Opus 4.7's confidence interval is greater than the score itself. By comparison, Humanity's Last Exam has 2500 problems \cite{hle-paper} and GAIA has 466 problems \cite{gaia-paper}.}
\label{fig:relaybench-wilson-ci}
\end{figure}

Many \relaybench\ problems are unsolvable without web search or internalization of esoteric knowledge extracted from the web. Additionally, some subproblems are impractical to solve without code-execution (\textit{e.g.}, solving ten sudoku games). Search problems intentionally rely on ground truth sources that are unlikely to change. However, if an answer becomes unavailable, it requires replacement, preventing \relaybench\ from being maintenance free indefinitely. In the future, a similar benchmark narrowing the scope of capabilities required to reach a solution, particularly one that removes search and code execution subproblems, may be a better measure of LLM capabilities. Alternatively, code execution and search tools could be sandboxed, although such restrictions would sacrifice the simplicity and reproducibility of the evaluation framework. These restrictions were not applied to \relaybench\ to avoid reducing the already small problem set. Models' low scores on \relaybench\ combined with the small problem count resulted in large confidence intervals for model performance on this benchmark (\figref{fig:relaybench-wilson-ci}).

Some subproblems in the test set require either knowledge of---or the ability to search for and process---documents within a corpus where the correct document meets a set of criteria that are non-trivial to evaluate. Solving these subproblems may require a new approach to managing context windows given that frontier LLMs cannot fit all the documents in the search space in their current context windows, which are presently limited to approximately one million tokens. These problems would be especially time-consuming for humans attempting to solve them compared to the rest of the test set. It is a common practice to compare LLMs to human panels, but doing so is neither essential to differentiate frontier models nor achievable given the limited budget and scope of this research. However, evaluating humans on \relaybench\ problems would provide insight into the duration it takes humans to solve these problems, and by extension determine how much of a human's time can be replicated with a one-shot request to an LLM.

Substituting the bespoke harness developed to evaluate this benchmark with a robust harness like Stirrup~\cite{stirrup-repo}, which is presently used to run several benchmarks in the Artificial Analysis Intelligence Index~\cite{aa-intelligence-methodology} and has seen adoption in model benchmarking~\cite{assetopsbench,nemotron3-ultra}, may improve the accuracy of this evaluation. Claude Opus 4.7 experienced far more execution errors than the other two models and, if its ratio of correct to parseable answers were maintained across all problems, would lead in accuracy.

Based on the limitations of \relaybench{}, the foremost concerns of an effort to create a sharpened revision that supplants the original should be growing the problem set and replacing existing problems with substitutes requiring a narrower set of capabilities to reach a solution.

\clearpage
\onecolumn
\phantomsection
\section*{Appendix}
\label{sec:appendix}

\appendix
\renewcommand{\thesection}{\Alph{section}}
\renewcommand{\theHsection}{appendix.\Alph{section}}
\titleformat{\section}[hang]
  {\large\bfseries}
  {\thesection \ \ \ /}
  {0.75em}
  {\MakeUppercase}

\section{Public example problem}\label{sec:public-example-problem}

\begin{promptbox}
\begin{Verbatim}
A benchmark for evaluating language models on mathematics competition problems, MathArena, has a paper associated with it. The paper contains at least one, but possibly multiple ranking tables. One table appears directly above a performance graph that plots results for o4-mini and Qwen3-30B-A3B, among other models.

In that specific table, each row contains a label followed by additional information in parentheses. What is the string that appears immediately to the left of the first opening parenthesis in the 4th row of that table? Trim any leading or trailing spaces from your answer. Put your answer in all caps and replace any spaces with underscores.

Let A_STR be that string.

Example: "GEMINI_2.0_FLASH"

In graph theory, the local complementation of a graph G at a vertex v is the operation that toggles every edge between distinct neighbors of v, replacing the induced subgraph on the open neighborhood N(v) with its complement, while leaving all edges not between two neighbors of v unchanged.

A graph H is a vertex-minor of a graph G if H can be obtained from G by first taking an induced subgraph and then applying a finite sequence of local complementations.

For a positive integer k, define f(k) as the smallest integer n such that every simple graph on exactly n vertices contains the edgeless graph on k vertices (k pairwise non-adjacent vertices with no edges among them) as a vertex-minor.

It is known that f(k) = 2^k - 1 for k = 1, 2, and 3.

What is f(4)? Let B_STR be that integer.

Example: "7"

In a single bar of a song by Kendrick Lamar, the names of two different real people are merged together into one composite name. Identify both people and return their full legal names (including all given names and any middle names). List both people's full legal names in alphabetical order by last name, separated by a single underscore. Use an underscore anywhere you would use a space in their names. Put your answer in all caps.

Let C_STR be that string.

Example: "ABIGAIL_MAY_JONES_ZACHARY_BLAKE_SMITH"

What woman is implied to be a part of a Kendrick album, exists as a concept rather than a literal person, and uses a gun? Put your answer in all caps. Replace any spaces with underscores.

Let D_STR be that string.

Example: "SCARLET_WITCH"

A slot-table is a data structure with n slots supporting three operations:

- Allocate(key) -> ref: Claims a free slot for `key` and returns a compressed reference `ref`.
- Dereference(key, ref) -> slot_index: Returns the index of the slot claimed by `key`. Only valid when called with the same key that produced `ref`.
- Free(key, ref): Releases the slot.

Because `ref` is only meaningful together with its owner `key`, the pair (key, ref) encodes the slot index rather than `ref` alone. This allows `ref` to use far fewer bits than the log(n)-bit lower bound that would apply to a standard standalone pointer into an array of size n.

The table operates at load factor 1 - 1/k, meaning at most n(1 - 1/k) slots can be simultaneously occupied.

Fixed-size references require every `ref` to use the same number of bits. A 2021 theoretical computer science result established tight asymptotic bounds -- a matching construction and a lower bound -- on the minimum achievable fixed-size reference length.

What is the asymptotically tight bound in Theta-notation on the optimal fixed-size reference size in bits, as a function of n and k?

Answer format rules for E_STR (follow exactly or your answer will not match):
- Use the ASCII uppercase letter O (not "Theta" or "Big-Theta")
- Every log must have parentheses around its argument: log(x), never log x
- You do not need to specify the base of any log
- Exponents use the form x^(y) -- parentheses are non-optional
- If a term contains both a variable and a log of that variable, the variable goes to the left of the log: xlog(x), not log(x)x
- If your answer has multiple additive terms, order them left to right from most characters to fewest characters
- Spaces to the left and right of + and - symbols

Let E_STR be the Theta expression exactly as written.

Example: "O(nlog(n) + n^(2))"

The Alliant Cashback Visa Signature offered 2.5% flat-rate cashback on purchases. What is the exact date that the discontinuation of this perk took effect? Format your answer as YYYYMMDD with no dashes or other separators.

Let F_STR be that date string.

Example: "20240101"

Fill in the blanks:

_ _____ ____ __________

Each underscore represents one letter and spaces separate words. The answer is four words with 1, 5, 4, and 10 letters respectively.

The answer is something related to all of the following: cult, gun, YouTube, grapple hook, island, MatPat, claymation, zombies.

Put your answer in all caps and replace any spaces with underscores.

Let G_STR be that string.

Example: "THE_POWER_OF_FRIENDSHIP"

As an LLM, you are allowed to use any tools, skills, or other features at your disposal to solve this problem.

Put your answer in a code block like this:
```
A_STR-B_STR-C_STR-D_STR-E_STR-F_STR-G_STR
```

Example: "GEMINI_2.0_FLASH-723-ABIGAIL_MAY_JONES_ZACHARY_BLAKE_SMITH-SCARLET_WITCH-O(nlog(n) + n^(2))-20240101-THE_POWER_OF_FRIENDSHIP"

For string variables A, C, D, and G: any non all-caps letter should be replaced with an underscore. Include B_STR, E_STR, and F_STR exactly as computed with no modification.

Throughout this problem when using big-O, it means big-theta. No non-ASCII characters should appear in your answer.

Do not include any other code blocks in your response.
\end{Verbatim}
\end{promptbox}

\section{Public example problem (encoded form)}\label{sec:public-example-problem-encoded}

\begin{promptbox}
\begin{Verbatim}
<user_context>
  <preferences>
      `/uc` anywhere in prompt = ultra-concise mode. Respond with the full answer and zero opening/closing remarks, no unnecessary words. Examples: integral → "127". Auction list → bare bullets, no preamble. All reasoning, verification, and explanation go in `<thinking>` blocks only — never in the response. Think extensively in `<thinking>` to ensure correctness before the ultra-concise reply.

      BAD /uc response example (never do this): Q: "which auctions not strategy proof /uc" BAD A: "From the slides, mechanisms not strategy-proof: - English auction – strategy-proof only under private values - First-price sealed-bid – bidders shade bids - Dutch auction – equivalent to first-price" Wrong because: has preamble, inline explanations, conversational framing. GOOD A: "- First-price sealed-bid\n- Dutch auction"
  </preferences>
  <user_summary>
      **Work context**

      Ethan Marlowe is a student at a top-15 technical university in the northeastern US with part-time self-employment income from software contracting work. He is also working as a trainer/reviewer on an AI image evaluation and ranking project for a tech client.

      **Personal context**

      Ethan has interests in competitive programming, DC and indie comics, music (hip-hop, indie folk, art rock, hyperpop), and investing. He is part of a student engineering team at his university, involved in both the technical and documentation side of vehicle design. He collects character plushies and has interest in personal web aesthetics (retro web buttons, personal static hosting). Ethan uses a desktop all-in-one machine with a SoC (system on a chip) architecture.

      **Top of mind**

      Ethan is working on his engineering team's annual race event, including race time estimation across multiple competing teams using timing data. His team's current vehicle documentation involves technical copywriting and material science details including thermal properties of polycarbonate filament and high-strength aluminum alloy use.

      **Brief history**

      *Recent months*

      Ethan has been deeply involved in a university competitive programming competition (launched mid-March), building a Python bot with a sense-think-act architecture and goal-based voting system. Key work included a full vision radius sense module, conveyor chain logic, harvester placement, enemy infrastructure destruction via self-destruct, and fixing a series of import and movement bugs.

      Ethan completed his prior-year tax filings across federal, two state returns, and a local return (mailed with check). His home address is in a township outside a major city, meaning no local income tax obligation from the city itself. He filed a local return in his university city at a ~3% EIT rate for part-year residency and self-employment income.

      He opened a new checking account for a sign-up bonus and is pursuing multiple bank account bonuses simultaneously. He also opened a Roth IRA and a rewards credit card during a session exploring financial aid interactions with retirement accounts, and is building a stock portfolio with holdings including large-cap tech and fintech. His investment thesis centers on robust structural monopolies with consistent revenue growth.

      Ethan is building and maintaining a personal infrastructure system: a cloud object storage bucket, a serverless worker, and a GitHub repo for his personal note vault. He has an HTML upload tool for uploading encrypted audio and images, with fixes for binary corruption, special character handling, and web archiving. He also runs a shell script to batch-archive files to the Wayback Machine.

      Ethan is managing postural issues with a flexibility routine and has an active skincare routine involving multiple prescription and OTC products. He is on a dermatologist-prescribed treatment course affecting shaving and product choices.

      He took a linguistics exam covering phonetics, phonology, morphology, and syntax, and completed a homework assignment for a game theory/social choice course. He also studied bigram language models and Laplace smoothing for an applied machine learning quiz.

      *Earlier context*

      Ethan registered to vote in his university's city at his campus-adjacent address. He filed quarterly estimated taxes as a self-employed contractor. He explored credit card strategy, cashback optimization, and bank account bonuses. He researched leveraged index investing, P/E ratio analysis, and major AI lab funding structures.

      He developed an IPA drag-and-drop study app (self-contained HTML, dark academic aesthetic) and an interval timer HTML tool. He built and debugged a CSS snippet for link color customization in his note-taking app.

      *Long-term background*

      Ethan has sustained interests in AI model architecture, competitive programming infrastructure, personal finance optimization, and comic books. He has explored retro web aesthetics, local AI model execution, and has ongoing familiarity with serverless workers, GitHub Actions, and cloud sync tooling.
  </user_summary>
  <recent_changes>
      - Music preference: prefers concept albums where the artist plays a character and tells a story; especially values interludes and skits that contribute to worldbuilding over conventional songs.
      - Emoji display requests: use visualize:show_widget (never a file). Show max relevant emojis as cards (64px emoji, label, Copy button). Copy btn: always black via inline btn.style.cssText with !important on all states. On copy: color:#4ade80 "Copied!" for 1.5s. No hover bg changes. Cards: var(--bg-200) bg, var(--border-100) border, 14px radius.
      - User is a university student, born 2005.
      - User skills live on the user's machine at Desktop/Notes/Agent Skills/<skill-name>/SKILL.md. This is the single source of truth.
      - Skill edit workflow: (1) read SKILL.md, (2) edit for diffs, (3) copy SKILL.md AND every file in references/, (4) package into a zip, (5) present the .skill zip.
      - New skill workflow: create directory, write blank SKILL.md, paste full content, then sync to assistant environment via copy + present_files. Never write full content from assistant's side.
      - When asked to create a bookmark: create a .url file with content `[InternetShortcut]\nURL=notes://web-open?url=SITE_URL`, filename = bookmark title.
      - The @ folder stores two .url file types: bookmarks (notes://web-open?url=...) and aliases (notes://open?vault=Notes&file=NOTE_NAME).
      - For comparisons, prefer a table. When /uc is active, especially favor a table if possible.
      - When user references editing files without specifying a path, it's always on the Desktop or in the notes vault directory.
      - For carbon offset prices, always search online for current market rates. Never give a made-up range — cite a real source.
      - For diagrams: never use the visualize tool. Use an SVG artifact instead to reduce token usage.
      - Addresses: university address — 400 W. Garrison Ave, Apt 214, [City], [State] [ZIP]. Home address — 3318 Fenwick Ct, [City], [State] [ZIP]. Mailing address — student mailbox on campus.
  </recent_changes>
  <summary_of_last_chat>
      User was debugging a flaky CSS animation on his personal note vault landing page — a marquee of retro web buttons that would occasionally desync after the tab regained focus. We narrowed it down to a stale requestAnimationFrame handle held across visibilitychange events, swapped in a CSS-only keyframe approach for the steady-state loop, and kept JS only for the pause-on-hover behavior. He then asked for a small SVG divider in the same 88x31 button aesthetic to sit above the marquee.
  </summary_of_last_chat>
</user_context>

You are being challenged for a benchmark. The rest of this prompt has been encoded. Use the dictionary below to decode the prompt. Do not encode your final answer to the prompt below.

```
{
  <rCL> : neighbors
  <yWI> : POWER
  <Djk> : Claims
  <sPb> : distinct
  <ArI> : performance
  <eKu> : separators
  <avo> : last
  <OHv> : but
  <gOg> : caps
  <uVe> : number
  <psp> : has
  <xSZ> : from
  <SMb> : occupied
  <zcW> : followed
  <gVD> : be
  <cyr> : Spaces
  <axw> : specify
  <EKS> : mathematics
  <XMh> : standalone
  <pbv> : complement
  <Zmp> : result
  <ExV> : no
  <JJz> : leaving
  <mWT> : exactly
  <khZ> : two
  <phr> : expression
  <Efc> : exact
  <Asr> : fewest
  <KkM> : respectively
  <iHw> : took
  <nsB> : hook
  <GEv> : fewer
  <TfK> : Each
  <AFn> : blanks
  <BCG> : You
  <ZrT> : owner
  <oHp> : together
  <ndx> : MathArena
  <NGH> : List
  <LHZ> : term
  <Vnv> : asymptotically
  <PgQ> : Visa
  <iJY> : implied
  <xiW> : notation
  <uVw> : it
  <QwG> : What
  <EaR> : merged
  <gxz> : theta
  <yTY> : features
  <Tek> : most
  <JuZ> : standard
  <lYD> : ?
  <YXK> : >
  <XEE> : Fill
  <WYc> : Replace
  <sqL> : s
  <ACh> : operates
  <IZD> : FLASH
  <gmj> : free
  <oWc> : structure
  <gzB> : among
  <vMk> : skills
  <uKX> : smallest
  <oPz> : row
  <RNF> : load
  <mds> : No
  <oIO> : known
  <eOo> : ^
  <prh> : mini
  <dVt> : integer
  <OpO> : argument
  <LQz> : by
  <BmK> : MAY
  <IrP> : following
  <gxy> : -
  <BGZ> : finite
  <FJU> : allows
  <yBr> : G
  <rKT> : matching
  <VPu> : spaces
  <LLc> : results
  <cwu> : code
  <pRz> : THE
  <Qgf> : left
  <NYA> : to
  <NDA> : Use
  <pPL> : operation
  <KNh> : need
  <eNa> : 3
  <tiT> : 1
  <cIt> : names
  <Qvu> : Answer
  <KkX> : If
  <tEy> : offered
  <Kfk> : name
  <FuI> : vertices
  <TyO> : of
  <aXy> : including
  <xWC> : anywhere
  <SxB> : not
  <FVQ> : MatPat
  <DJz> : can
  <fUP> : In
  <MkU> : and
  <JVi> : other
  <TPx> : +
  <NuI> : one
  <NAO> : Returns
  <VBP> : four
  <nsH> : .
  <NtB> : graph
  <GyP> : literal
  <ONV> : induced
  <blh> : complementation
  <jLN> : bar
  <yfW> : theory
  <HSI> : %
  <Nnb> : Do
  <pmf> : Put
  <jfp> : when
  <RAy> : complementations
  <SXl> : response
  <oYW> : single
  <pGz> : with
  <QXS> : specific
  <dQN> : Only
  <iID> : bits
  <loP> : label
  <Kgy> : would
  <FJH> : big
  <HSs> : underscores
  <WUk> : Dereference
  <gxb> : size
  <vSu> : GEMINI
  <bcz> : '
  <Cbr> : disposal
  <DuY> : modification
  <vMH> : into
  <DVW> : WITCH
  <Gea> : such
  <tqN> : then
  <ZRs> : means
  <pQB> : solve
  <KFi> : benchmark
  <PXQ> : its
  <EyO> : are
  <UbW> : data
  <eVK> : lower
  <PzH> : additional
  <vta> : neighborhood
  <RCG> : is
  <MhQ> : any
  <RHy> : Signature
  <vBt> : Unicode
  <MWr> : leading
  <pUM> : use
  <APy> : YouTube
  <NCG> : edgeless
  <KdF> : reference
  <KQJ> : non
  <uMR> : or
  <HKo> : xlog
  <JYw> : log
  <BEI> : full
  <MAv> : k
  <pzd> : Example
  <mRY> : index
  <Exu> : local
  <qyv> : your
  <kzX> : claimed
  <ZEv> : replacing
  <zaL> : =
  <Wny> : taking
  <YZh> : order
  <gUq> : should
  <mIH> : each
  <TtM> : slot
  <aaa> : words
  <MOA> : apply
  <sAJ> : N
  <qkv> : Cashback
  <AgK> : subgraph
  <Cjx> : something
  <UyH> : key
  <iQH> : toggles
  <ICk> : parentheses
  <jOr> : ref
  <uol> : pairwise
  <lrV> : Throughout
  <rlx> : in
  <wAY> : legal
  <FsZ> : the
  <IVg> : simple
  <Ydj> : between
  <cMt> : minimum
  <SqB> : dashes
  <wPo> : o
  <Let> : string
  <bRh> : island
  <QXh> : Allocate
  <Bye> : This
  <HLO> : three
  <gba> : alone
  <UTe> : rate
  <IvY> : separate
  <NKz> : SCARLET
  <XYo> : O
  <xNY> : minor
  <QGU> : competition
  <iKT> : ABIGAIL
  <fFy> : woman
  <imI> : adjacent
  <mCH> : grapple
  <CsO> : evaluating
  <Uqw> : SMITH
  <xma> : C
  <CKl> : trailing
  <eAd> : opening
  <GbF> : form
  <OAl> : while
  <oOn> : STR
  <TLY> : blocks
  <geg> : compressed
  <uNb> : Fixed
  <RrM> : symbols
  <IWm> : asymptotic
  <bBV> : a
  <Juw> : cashback
  <NVz> : The
  <cVn> : alphabetical
  <HvW> : them
  <aWm> : if
  <NuY> : f
  <Ezq> : flat
  <UGi> : middle
  <QRw> : problems
  <qbO> : replace
  <zJB> : do
  <fJN> : real
  <YUh> : models
  <IYT> : Exponents
  <iZC> : function
  <MGS> : Alliant
  <kJh> : only
  <KFn> : 7
  <giX> : `
  <EVn> : computer
  <MyM> : BLAKE
  <zJH> : :
  <eNb> : established
  <YGN> : like
  <vuQ> : OF
  <mqp> : One
  <PaC> : /
  <TAl> : language
  <nat> : tables
  <RsY> : returns
  <sSU> : open
  <dJg> : Big
  <dph> : (
  <GFv> : both
  <ptB> : for
  <BFP> : supporting
  <gjf> : Let
  <cCu> : F
  <TeL> : space
  <aLs> : problem
  <vBU> : related
  <WEU> : given
  <bNc> : information
  <DNh> : base
  <dCh> : first
  <RmM> : different
  <Zgt> : person
  <tqc> : ,
  <rxl> : album
  <kMm> : when
  <CiP> : their
  <UvZ> : immediately
  <huu> : 0
  <qhY> : FRIENDSHIP
  <WiY> : Identify
  <OEo> : array
  <Dmi> : Lamar
  <UtQ> : represents
  <oVz> : length
  <dzf> : least
  <Uqo> : For
  <acI> : this
  <jLk> : rather
  <oVM> : table
  <Uee> : edge
  <Wyo> : possibly
  <Xdq> : Include
  <soY> : return
  <OgM> : achievable
  <RwJ> : exists
  <xpP> : far
  <Fam> : "
  <tUs> : allowed
  <PKI> : slots
  <hTv> : must
  <LCV> : character
  <uoI> : produced
  <StI> : H
  <DTM> : optimal
  <vRQ> : variable
  <aCx> : perk
  <QKU> : Kendrick
  <cGa> : tools
  <ICY> : associated
  <nyv> : Qwen
  <Oou> : 5
  <xNw> : Free
  <pLA> : ranking
  <Tor> : written
  <Qzc> : tight
  <YjI> : format
  <Fiu> : effect
  <ERW> : letters
  <zgn> : date
  <lTA> : fixed
  <Isj> : concept
  <ShZ> : called
  <QFo> : gun
  <GGe> : define
  <TVi> : require
  <Bcf> : _
  <KHO> : variables
  <BEO> : th
  <wUr> : every
  <GOm> : same
  <fmU> : applying
  <KXI> : obtained
  <din> : include
  <oHZ> : 4
  <QIb> : at
  <jcm> : goes
  <VSZ> : JONES
  <Qax> : never
  <pvy> : meaning
  <cLv> : bit
  <cxi> : zombies
  <Wki> : operations
  <ZxL> : plots
  <SIk> : letter
  <UyT> : pair
  <gsj> : parenthesis
  <yPU> : computed
  <SVV> : claymation
  <lHq> : D
  <Cqd> : on
  <VaB> : E
  <rdm> : characters
  <tBO> : uses
  <OhH> : using
  <diL> : Format
  <Rjf> : cult
  <BFL> : around
  <TqF> : meaningful
  <kLB> : It
  <xfF> : bounds
  <RTV> : v
  <Icm> : follow
  <uSp> : song
  <nGQ> : that
  <iFh> : encodes
  <HQX> : as
  <FRD> : references
  <Wkl> : than
  <Bti> : answer
  <TgU> : match
  <QMt> : contains
  <yXQ> : you
  <IqZ> : right
  <ctS> : simultaneously
  <MhO> : appear
  <oaf> : part
  <MZy> : Trim
  <atx> : y
  <Iwb> : positive
  <gvZ> : A
  <IOe> : multiple
  <mYI> : purchases
  <uZj> : block
  <Ods> : LLM
  <Ctn> : have
  <vvS> : paper
  <RHH> : separated
  <YfC> : valid
  <LVD> : people
  <Qdp> : terms
  <POy> : YYYYMMDD
  <Umn> : ZACHARY
  <Llz> : ASCII
  <YTt> : unchanged
  <lNZ> : nlog
  <ufF> : additive
  <Kws> : n
  <yGF> : directly
  <zAC> : Releases
  <NDo> : bound
  <sey> : Theta
  <ajQ> : theoretical
  <ZUw> : x
  <kkG> : composite
  <cOS> : factor
  <BGb> : sequence
  <jQO> : construction
  <auW> : B
  <ETe> : discontinuation
  <foF> : all
  <YNh> : Because
  <ebp> : 2
  <jcU> : underscore
  <BQY> : rules
  <jzP> : )
  <mYs> : appears
  <yuL> : above
  <udh> : an
  <SMe> : pointer
  <iDq> : vertex
  <hPR> : science
  <YJu> : will
  <sIC> : optional
  <Umv> : replaced
  <ZOV> : As
  <xut> : edges
  <xdv> : Every
}
```

Do what that prompt says and follow its instructions. The prompt is below:

<gvZ> <KFi> <ptB> <CsO> <TAl> <YUh> <Cqd> <EKS> <QGU> <QRw><tqc> <ndx><tqc> <psp> <bBV> <vvS> <ICY> <pGz> <uVw><nsH> <NVz> <vvS> <QMt> <QIb> <dzf> <NuI><tqc> <OHv> <Wyo> <IOe> <pLA> <nat><nsH> <mqp> <oVM> <mYs> <yGF> <yuL> <bBV> <ArI> <NtB> <nGQ> <ZxL> <LLc> <ptB> <wPo><oHZ><gxy><prh> <MkU> <nyv><eNa><gxy><eNa><huu><auW><gxy><gvZ><eNa><auW><tqc> <gzB> <JVi> <YUh><nsH>

<fUP> <nGQ> <QXS> <oVM><tqc> <mIH> <oPz> <QMt> <bBV> <loP> <zcW> <LQz> <PzH> <bNc> <rlx> <ICk><nsH> <QwG> <RCG> <FsZ> <Let> <nGQ> <mYs> <UvZ> <NYA> <FsZ> <Qgf> <TyO> <FsZ> <dCh> <eAd> <gsj> <rlx> <FsZ> <oHZ><BEO> <oPz> <TyO> <nGQ> <oVM><lYD> <MZy> <MhQ> <MWr> <uMR> <CKl> <VPu> <xSZ> <qyv> <Bti><nsH> <pmf> <qyv> <Bti> <rlx> <foF> <gOg> <MkU> <qbO> <MhQ> <VPu> <pGz> <HSs><nsH>

<gjf> <gvZ><Bcf><oOn> <gVD> <nGQ> <Let><nsH>

<pzd><zJH> <Fam><vSu><Bcf><ebp><nsH><huu><Bcf><IZD><Fam>

<fUP> <NtB> <yfW><tqc> <FsZ> <Exu> <blh> <TyO> <bBV> <NtB> <yBr> <QIb> <bBV> <iDq> <RTV> <RCG> <FsZ> <pPL> <nGQ> <iQH> <wUr> <Uee> <Ydj> <sPb> <rCL> <TyO> <RTV><tqc> <ZEv> <FsZ> <ONV> <AgK> <Cqd> <FsZ> <sSU> <vta> <sAJ><dph><RTV><jzP> <pGz> <PXQ> <pbv><tqc> <OAl> <JJz> <foF> <xut> <SxB> <Ydj> <khZ> <rCL> <TyO> <RTV> <YTt><nsH>

<gvZ> <NtB> <StI> <RCG> <bBV> <iDq><gxy><xNY> <TyO> <bBV> <NtB> <yBr> <aWm> <StI> <DJz> <gVD> <KXI> <xSZ> <yBr> <LQz> <dCh> <Wny> <udh> <ONV> <AgK> <MkU> <tqN> <fmU> <bBV> <BGZ> <BGb> <TyO> <Exu> <RAy><nsH>

<Uqo> <bBV> <Iwb> <dVt> <MAv><tqc> <GGe> <NuY><dph><MAv><jzP> <HQX> <FsZ> <uKX> <dVt> <Kws> <Gea> <nGQ> <wUr> <IVg> <NtB> <Cqd> <mWT> <Kws> <FuI> <QMt> <FsZ> <NCG> <NtB> <Cqd> <MAv> <FuI> <dph><MAv> <uol> <KQJ><gxy><imI> <FuI> <pGz> <ExV> <xut> <gzB> <HvW><jzP> <HQX> <bBV> <iDq><gxy><xNY><nsH>

<kLB> <RCG> <oIO> <nGQ> <NuY><dph><MAv><jzP> <zaL> <ebp><eOo><MAv> <gxy> <tiT> <ptB> <MAv> <zaL> <tiT><tqc> <ebp><tqc> <MkU> <eNa><nsH>

<QwG> <RCG> <NuY><dph><oHZ><jzP><lYD> <gjf> <auW><Bcf><oOn> <gVD> <nGQ> <dVt><nsH>

<pzd><zJH> <Fam><KFn><Fam>

<fUP> <bBV> <oYW> <jLN> <TyO> <bBV> <uSp> <LQz> <QKU> <Dmi><tqc> <FsZ> <cIt> <TyO> <khZ> <RmM> <fJN> <LVD> <EyO> <EaR> <oHp> <vMH> <NuI> <kkG> <Kfk><nsH> <WiY> <GFv> <LVD> <MkU> <soY> <CiP> <BEI> <wAY> <cIt> <dph><aXy> <foF> <WEU> <cIt> <MkU> <MhQ> <UGi> <cIt><jzP><nsH> <NGH> <GFv> <LVD><bcz><sqL> <BEI> <wAY> <cIt> <rlx> <cVn> <YZh> <LQz> <avo> <Kfk><tqc> <RHH> <LQz> <bBV> <oYW> <jcU><nsH> <NDA> <udh> <jcU> <xWC> <yXQ> <Kgy> <pUM> <bBV> <TeL> <rlx> <CiP> <cIt><nsH> <pmf> <qyv> <Bti> <rlx> <foF> <gOg><nsH>

<gjf> <xma><Bcf><oOn> <gVD> <nGQ> <Let><nsH>

<pzd><zJH> <Fam><iKT><Bcf><BmK><Bcf><VSZ><Bcf><Umn><Bcf><MyM><Bcf><Uqw><Fam>

<QwG> <fFy> <RCG> <iJY> <NYA> <gVD> <bBV> <oaf> <TyO> <bBV> <QKU> <rxl><tqc> <RwJ> <HQX> <bBV> <Isj> <jLk> <Wkl> <bBV> <GyP> <Zgt><tqc> <MkU> <tBO> <bBV> <QFo><lYD> <pmf> <qyv> <Bti> <rlx> <foF> <gOg><nsH> <WYc> <MhQ> <VPu> <pGz> <HSs><nsH>

<gjf> <lHq><Bcf><oOn> <gVD> <nGQ> <Let><nsH>

<pzd><zJH> <Fam><NKz><Bcf><DVW><Fam>

<gvZ> <TtM><gxy><oVM> <RCG> <bBV> <UbW> <oWc> <pGz> <Kws> <PKI> <BFP> <HLO> <Wki><zJH>

<gxy> <QXh><dph><UyH><jzP> <gxy><YXK> <jOr><zJH> <Djk> <bBV> <gmj> <TtM> <ptB> <giX><UyH><giX> <MkU> <RsY> <bBV> <geg> <KdF> <giX><jOr><giX><nsH>
<gxy> <WUk><dph><UyH><tqc> <jOr><jzP> <gxy><YXK> <TtM><Bcf><mRY><zJH> <NAO> <FsZ> <mRY> <TyO> <FsZ> <TtM> <kzX> <LQz> <giX><UyH><giX><nsH> <dQN> <YfC> <jfp> <ShZ> <pGz> <FsZ> <GOm> <UyH> <nGQ> <uoI> <giX><jOr><giX><nsH>
<gxy> <xNw><dph><UyH><tqc> <jOr><jzP><zJH> <zAC> <FsZ> <TtM><nsH>

<YNh> <giX><jOr><giX> <RCG> <kJh> <TqF> <oHp> <pGz> <PXQ> <ZrT> <giX><UyH><giX><tqc> <FsZ> <UyT> <dph><UyH><tqc> <jOr><jzP> <iFh> <FsZ> <TtM> <mRY> <jLk> <Wkl> <giX><jOr><giX> <gba><nsH> <Bye> <FJU> <giX><jOr><giX> <NYA> <pUM> <xpP> <GEv> <iID> <Wkl> <FsZ> <JYw><dph><Kws><jzP><gxy><cLv> <eVK> <NDo> <nGQ> <Kgy> <MOA> <NYA> <bBV> <JuZ> <XMh> <SMe> <vMH> <udh> <OEo> <TyO> <gxb> <Kws><nsH>

<NVz> <oVM> <ACh> <QIb> <RNF> <cOS> <tiT> <gxy> <tiT><PaC><MAv><tqc> <pvy> <QIb> <Tek> <Kws><dph><tiT> <gxy> <tiT><PaC><MAv><jzP> <PKI> <DJz> <gVD> <ctS> <SMb><nsH>

<uNb><gxy><gxb> <FRD> <TVi> <wUr> <giX><jOr><giX> <NYA> <pUM> <FsZ> <GOm> <uVe> <TyO> <iID><nsH> <gvZ> <ebp><huu><ebp><tiT> <ajQ> <EVn> <hPR> <Zmp> <eNb> <Qzc> <IWm> <xfF> <gxy><gxy> <bBV> <rKT> <jQO> <MkU> <bBV> <eVK> <NDo> <gxy><gxy> <Cqd> <FsZ> <cMt> <OgM> <lTA><gxy><gxb> <KdF> <oVz><nsH>

<QwG> <RCG> <FsZ> <Vnv> <Qzc> <NDo> <rlx> <sey><gxy><xiW> <Cqd> <FsZ> <DTM> <lTA><gxy><gxb> <KdF> <gxb> <rlx> <iID><tqc> <HQX> <bBV> <iZC> <TyO> <Kws> <MkU> <MAv><lYD>

<Qvu> <YjI> <BQY> <ptB> <VaB><Bcf><oOn> <dph><Icm> <mWT> <uMR> <qyv> <Bti> <YJu> <SxB> <TgU><jzP><zJH>
<gxy> <NDA> <FsZ> <vBt> <LCV> <XYo> <dph><SxB> <Fam><sey><Fam> <uMR> <Fam><dJg><gxy><sey><Fam><jzP>
<gxy> <xdv> <JYw> <hTv> <Ctn> <ICk> <BFL> <PXQ> <OpO><zJH> <JYw><dph><ZUw><jzP><tqc> <Qax> <JYw> <ZUw>
<gxy> <BCG> <zJB> <SxB> <KNh> <NYA> <axw> <FsZ> <DNh> <TyO> <MhQ> <JYw>
<gxy> <IYT> <pUM> <FsZ> <GbF> <ZUw><eOo><dph><atx><jzP> <gxy><gxy> <ICk> <EyO> <KQJ><gxy><sIC>
<gxy> <KkX> <bBV> <LHZ> <QMt> <GFv> <bBV> <vRQ> <MkU> <bBV> <JYw> <TyO> <nGQ> <vRQ><tqc> <FsZ> <vRQ> <jcm> <NYA> <FsZ> <Qgf> <TyO> <FsZ> <JYw><zJH> <HKo><dph><ZUw><jzP><tqc> <SxB> <JYw><dph><ZUw><jzP><ZUw>
<gxy> <KkX> <qyv> <Bti> <psp> <IOe> <ufF> <Qdp><tqc> <YZh> <HvW> <Qgf> <NYA> <IqZ> <xSZ> <Tek> <rdm> <NYA> <Asr> <rdm>
<gxy> <cyr> <NYA> <FsZ> <Qgf> <MkU> <IqZ> <TyO> <TPx> <MkU> <gxy> <RrM>

<gjf> <VaB><Bcf><oOn> <gVD> <FsZ> <sey> <phr> <mWT> <HQX> <Tor><nsH>

<pzd><zJH> <Fam><XYo><dph><lNZ><dph><Kws><jzP> <TPx> <Kws><eOo><dph><ebp><jzP><jzP><Fam>

<NVz> <MGS> <qkv> <PgQ> <RHy> <tEy> <ebp><nsH><Oou><HSI> <Ezq><gxy><UTe> <Juw> <Cqd> <mYI><nsH> <QwG> <RCG> <FsZ> <Efc> <zgn> <nGQ> <FsZ> <ETe> <TyO> <acI> <aCx> <iHw> <Fiu><lYD> <diL> <qyv> <Bti> <HQX> <POy> <pGz> <ExV> <SqB> <uMR> <JVi> <eKu><nsH>

<gjf> <cCu><Bcf><oOn> <gVD> <nGQ> <zgn> <Let><nsH>

<pzd><zJH> <Fam><ebp><huu><ebp><oHZ><huu><tiT><huu><tiT><Fam>

<XEE> <rlx> <FsZ> <AFn><zJH>

<Bcf> <Bcf><Bcf><Bcf><Bcf><Bcf> <Bcf><Bcf><Bcf><Bcf> <Bcf><Bcf><Bcf><Bcf><Bcf><Bcf><Bcf><Bcf><Bcf><Bcf>

<TfK> <jcU> <UtQ> <NuI> <SIk> <MkU> <VPu> <IvY> <aaa><nsH> <NVz> <Bti> <RCG> <VBP> <aaa> <pGz> <tiT><tqc> <Oou><tqc> <oHZ><tqc> <MkU> <tiT><huu> <ERW> <KkM><nsH>

<NVz> <Bti> <RCG> <Cjx> <vBU> <NYA> <foF> <TyO> <FsZ> <IrP><zJH> <Rjf><tqc> <QFo><tqc> <APy><tqc> <mCH> <nsB><tqc> <bRh><tqc> <FVQ><tqc> <SVV><tqc> <cxi><nsH>

<pmf> <qyv> <Bti> <rlx> <foF> <gOg> <MkU> <qbO> <MhQ> <VPu> <pGz> <HSs><nsH>

<gjf> <yBr><Bcf><oOn> <gVD> <nGQ> <Let><nsH>

<pzd><zJH> <Fam><pRz><Bcf><yWI><Bcf><vuQ><Bcf><qhY><Fam>

<ZOV> <udh> <Ods><tqc> <yXQ> <EyO> <tUs> <NYA> <pUM> <MhQ> <cGa><tqc> <vMk><tqc> <uMR> <JVi> <yTY> <QIb> <qyv> <Cbr> <NYA> <pQB> <acI> <aLs><nsH>

<pmf> <qyv> <Bti> <rlx> <bBV> <cwu> <uZj> <YGN> <acI><zJH>
<giX><giX><giX>
<gvZ><Bcf><oOn><gxy><auW><Bcf><oOn><gxy><xma><Bcf><oOn><gxy><lHq><Bcf><oOn><gxy><VaB><Bcf><oOn><gxy><cCu><Bcf><oOn><gxy><yBr><Bcf><oOn>
<giX><giX><giX>

<pzd><zJH> <Fam><vSu><Bcf><ebp><nsH><huu><Bcf><IZD><gxy><KFn><ebp><eNa><gxy><iKT><Bcf><BmK><Bcf><VSZ><Bcf><Umn><Bcf><MyM><Bcf><Uqw><gxy><NKz><Bcf><DVW><gxy><XYo><dph><lNZ><dph><Kws><jzP> <TPx> <Kws><eOo><dph><ebp><jzP><jzP><gxy><ebp><huu><ebp><oHZ><huu><tiT><huu><tiT><gxy><pRz><Bcf><yWI><Bcf><vuQ><Bcf><qhY><Fam>

<Uqo> <Let> <KHO> <gvZ><tqc> <xma><tqc> <lHq><tqc> <MkU> <yBr><zJH> <MhQ> <KQJ> <foF><gxy><gOg> <SIk> <gUq> <gVD> <Umv> <pGz> <udh> <jcU><nsH> <Xdq> <auW><Bcf><oOn><tqc> <VaB><Bcf><oOn><tqc> <MkU> <cCu><Bcf><oOn> <mWT> <HQX> <yPU> <pGz> <ExV> <DuY><nsH>

<lrV> <acI> <aLs> <kMm> <OhH> <FJH><gxy><XYo><tqc> <uVw> <ZRs> <FJH><gxy><gxz><nsH> <mds> <KQJ><gxy><Llz> <rdm> <gUq> <MhO> <rlx> <qyv> <Bti><nsH>

<Nnb> <SxB> <din> <MhQ> <JVi> <cwu> <TLY> <rlx> <qyv> <SXl><nsH>
\end{Verbatim}
\end{promptbox}

\end{document}